\documentclass[11pt]{article}
\usepackage{eacl2009}
\usepackage{times}
\usepackage{url}
\usepackage{latexsym}
\usepackage{graphicx}

\title{Discovering Global Patterns in Linguistic Networks through\\ Spectral Analysis: A Case Study of the Consonant Inventories}

\author{
Animesh Mukherjee\thanks{This research has been conducted during the author's internship at Microsoft Research India.}\\
  Indian Institute of Technology, Kharagpur\\
  {\tt animeshm@cse.iitkgp.ernet.in}
\AND
Monojit Choudhury \and Ravi Kannan\\
 Microsoft Research India \\
{\tt \{monojitc,kannan\}@microsoft.com}  }

\date{}

\begin{document}
\maketitle
\begin{abstract}
Recent research has shown that language and the socio-cognitive phenomena associated with it can be aptly modeled and 
visualized through networks of linguistic entities. However, most of the existing works on linguistic networks focus 
only on the local properties of the networks. This study is an attempt to analyze the structure of languages via a 
purely structural technique, namely spectral analysis, which is ideally suited for discovering the global 
correlations in a network. Application of this technique to PhoNet, the co-occurrence network of consonants, not only 
reveals several natural linguistic principles governing the structure of the consonant inventories, but is also able 
to quantify their relative importance. We believe that this powerful technique can be successfully applied, in general, to study the structure of natural languages.
\end{abstract}

\section{Introduction}
\label{intro}
Language and the associated socio-cognitive phenomena can be modeled as networks, where the nodes correspond to 
linguistic entities and the edges denote the pairwise interaction or relationship between these entities. The study 
of linguistic networks has been quite popular in the recent times and has provided us with several interesting 
insights into the nature of language (see Choudhury and Mukherjee~\shortcite{choudhury:08} for an extensive survey). Examples include study of the WordNet~\cite{sigman:02}, syntactic dependency network of words~\cite{cancho:05} and network of co-occurrence of consonants in sound inventories~\cite{coling:08,Mukherjee:06}. 

Most of the existing studies on linguistic networks, however, focus only on the local structural properties such as 
the degree and clustering coefficient of the nodes, and shortest paths between pairs of nodes. On the other hand, 
although it is a well known fact that the {\em spectrum} of a network can provide important information about 
its global structure, the use of this powerful mathematical machinery to infer global patterns in linguistic networks 
is rarely found in the literature. Note that spectral analysis, however, has been successfully employed in the domains 
of biological and social networks~\cite{farkas:01,mihail:03,Anirban:07}. In the context of linguistic networks, \cite{belkin-goldsmith:2002} is the only work we are aware of that analyzes the eigenvectors to obtain a two dimensional visualize of the network. Nevertheless, the work does not study the spectrum of the graph.  

The aim of the present work is to demonstrate the use of spectral analysis for discovering the global patterns in linguistic networks. These patterns, in turn, are then interpreted in the light of existing linguistic theories to gather deeper insights into the nature of the underlying linguistic phenomena.  We apply this rather generic technique to find the principles that are responsible for shaping the consonant inventories, which is a well researched problem in phonology since 1931~\cite{Trub:31,Lindblom:88,Boersma:98,Clements:04}.  The analysis is carried out on a network defined in~\cite{Mukherjee:06}, where the consonants are the nodes and there is an edge between two nodes $u$ and $v$ if the consonants corresponding to them co-occur in a language. The number of times they co-occur across languages define the weight of the edge. We explain the results obtained from the spectral analysis of the network post-facto using three linguistic principles. The method also automatically reveals the quantitative importance of each of these principles. 

It is worth mentioning here that earlier researchers have also noted the importance of the aforementioned principles. However, what was not known was how much importance one should associate with each of these principles. We also note that the technique of spectral analysis neither explicitly nor implicitly assumes that these principles exist or are important, but deduces them automatically. Thus, we believe that spectral analysis is a promising approach that is well suited to the discovery of linguistic principles underlying a set of observations represented as a network of entities. The fact that the principles ``discovered" in this study are already well established results adds to the credibility of the method. Spectral analysis of large linguistic networks in the future can possibly reveal hitherto unknown universal principles.

The rest of the paper is organized as follows. Sec.~\ref{sec:SA} introduces the technique of spectral analysis 
of networks and illustrates some of its applications. The problem of consonant inventories and how it can be modeled 
and studied within the framework of linguistic networks are described in Sec.~\ref{sec:phonet}. Sec.~\ref{sec:ESAPh} 
presents the spectral analysis of the consonant co-occurrence network, the observations and interpretations. 
Sec.~\ref{sec:disc} concludes by summarizing the work and the contributions and listing out future research 
directions.

\begin{figure*}
\begin{center}
\includegraphics[width=4in]{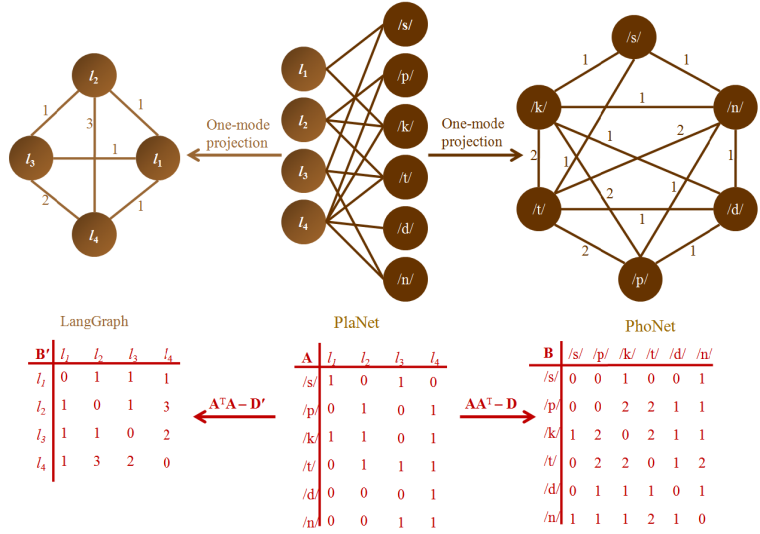}
\caption{Illustration of the nodes and edges of PlaNet and PhoNet along with their respective adjacency matrix 
representations.}\label{fig1}
\end{center}
\end{figure*}

\section{A Primer to Spectral Analysis}\label{sec:SA}

{\em Spectral analysis}\footnote{The term spectral analysis is also used in the context of signal processing, where 
it refers to the study of the frequency spectrum of a signal.} is a powerful tool capable of revealing the global structural patterns underlying an 
enormous and complicated environment of interacting entities. Essentially, it refers to the systematic study of the 
eigenvalues and the eigenvectors of the adjacency matrix of the network of these interacting entities. Here we shall 
briefly review the basic concepts involved in spectral analysis and describe some of its applications (see~\cite{chung:94,kannan:08} for details).

A network or a graph consisting of $n$ nodes (labeled as 1 through $n$) can be represented by a $n\times n$ square 
matrix {\bf A}, where the entry $a_{ij}$ represents the weight of the edge from node $i$ to node $j$. {\bf A}, which 
is known as the {\em adjacency matrix}, is symmetric for an undirected graph and have binary entries for an 
unweighted graph. $\lambda$ is an eigenvalue of {\bf A} if there is an $n$-dimensional vector $\mathbf{x}$ such that
\begin{displaymath}
\mathbf{Ax}=\lambda\mathbf{x}
\end{displaymath}
Any real symmetric matrix {\bf A} has $n$ (possibly non-distinct) eigenvalues $\lambda_0 \le \lambda_1 \le \dots \le 
\lambda_{n-1}$, and corresponding $n$ eigenvectors that are mutually orthogonal. The {\em spectrum} of a graph is the set of the distinct eigenvalues of the graph and their corresponding multiplicities. It is usually represented as a plot with the eigenvalues in x-axis and their 
multiplicities plotted in the y-axis. 

The spectrum of real and random graphs display several interesting properties. Banerjee and 
Jost~\shortcite{Anirban:07} report the spectrum of several biological networks that are significantly different from 
the spectrum of artificially generated graphs\footnote{Banerjee and Jost~\shortcite{Anirban:07} report the spectrum of the 
graph's Laplacian matrix rather than the adjacency matrix. It is increasingly popular these days to analyze the 
spectral properties of the graph's Laplacian matrix. However, for reasons explained later, here we will be conduct spectral analysis of the adjacency matrix rather than its Laplacian.}. Spectral analysis is also closely related to {\em Principal Component 
Analysis} and {Multidimensional Scaling}. If the first few (say $d$) eigenvalues of a matrix are much higher than the rest 
of the eigenvalues, then it can be concluded that the rows of the matrix can be approximately represented as linear 
combinations of $d$ orthogonal vectors. This further implies that the corresponding graph has a few motifs 
(subgraphs) that are repeated a large number of time to obtain the global structure of the graph~\cite{Anirban:08}. 

Spectral properties are representative of an n-dimensional average behavior of the 
underlying system, thereby providing considerable insight into its global organization. For example, the principal 
eigenvector (i.e., the eigenvector corresponding to the largest eigenvalue) is the direction in which the sum of the 
square of the projections of the row vectors of the matrix is maximum. In fact, the principal eigenvector of a graph 
is used to compute the centrality of the nodes, which is also known as {\em PageRank} in the context of WWW. 
Similarly, the second eigen vector component is used for graph clustering. 

In the next two sections we describe how spectral analysis can be applied to discover the organizing principles 
underneath the structure of consonant inventories.

\section{Consonant Co-occurrence Network}\label{sec:phonet}

The most basic unit of human languages are the speech sounds. The repertoire of sounds that make up the sound 
inventory of a language are not chosen arbitrarily even though the speakers are capable of producing and perceiving a 
plethora of them. In contrast, these inventories show exceptionally regular patterns across the languages of the 
world, which is in fact, a common point of consensus in phonology. Right from the beginning of the 20$^\textrm{th}$ 
century, there have been a large number of linguistically motivated attempts~\cite{Trub:39,Lindblom:88,Boersma:98,Clements:04} to explain the formation of these patterns across the consonant inventories. More recently, Mukherjee and his colleagues~\cite{acl:06,Mukherjee:06,coling:08} studied this problem in the framework of complex networks. Since here we shall conduct a spectral analysis of the network defined in Mukherjee et al.~\shortcite{Mukherjee:06}, we briefly survey the models and the important results of their work.

Choudhury et al.~\shortcite{acl:06} introduced a bipartite network model for the consonant inventories. Formally, a set of consonant inventories is represented as a graph $G = \langle V_L, V_C, E_{lc}\rangle$, where the nodes in 
one partition correspond to the languages ($V_L$) and that in the other partition correspond to the consonants 
($V_C$). There is an edge ($v_l$, $v_c$) between a language node $v_l \in V_L$ (representing the language $l$) and a 
consonant node $v_c \in V_C$ (representing the consonant $c$) iff the consonant $c$ is present in the inventory of 
the language $l$. This network is called the {\bf P}honeme-{\bf La}nguage {\bf Net}work or {\bf PlaNet} and represent the 
connections between the language and the consonant nodes through a 0-1 matrix $\mathbf{A}$ as shown by a hypothetical 
example in Fig.~\ref{fig1}. Further, in~\cite{Mukherjee:06}, the authors define the {\bf Pho}neme-Phoneme {\bf Net}work or {\bf PhoNet} as the one-mode 
projection of PlaNet onto the consonant nodes, i.e., a network $G= \langle V_C, E_{cc^{'}}\rangle$, where the nodes 
are the consonants and two nodes $v_c$ and $v_{c^{'}}$ are linked by an edge with weight equal to the number of 
languages in which both $c$ and $c'$ occur together. In other words, PhoNet can be expressed as a matrix $\mathbf{B}$ (see 
Fig.~\ref{fig1}) such that $\mathbf{B} = \mathbf{AA}^\textrm{T} - \mathbf{D}$ where $\mathbf{D}$ is a diagonal matrix with its 
entries corresponding to the frequency of occurrence of the consonants. Similarly, we can also construct the one-mode projection of PlaNet 
onto the language nodes (which we shall refer to as the Language-Language Graph or LangGraph) can be expressed as $\mathbf{B}' = \mathbf{A}^\textrm{T}\mathbf{A} - \mathbf{D}'$, where $\mathbf{D}'$ is a diagonal matrix with its entries corresponding to the size of the consonant inventories for each language.
 
The matrix $\mathbf{A}$ and hence, $\mathbf{B}$ and $\mathbf{B}'$ have been constructed from the UCLA Phonological Segment Inventory Database 
(UPSID)~\cite{Maddieson:84} that hosts the consonant inventories of 317 languages with a total of 541 consonants found across them. Note that, UPSID uses articulatory features to describe the consonants and assumes these features to be binary-valued, which in turn implies that every consonant can be represented by a binary vector. Later on, we shall use this representation for our experiments.

By construction, we have $|V_L|$ = 317, $|V_C|$ = 541, $|E_{lc}|$ = 7022, and $|E_{cc'}|$ = 30412. Consequently, the 
order of the matrix $\mathbf{A}$ is 541 $\times$ 317 and that of the matrix $\mathbf{B}'$ is 541 $\times$ 541. It has been found that the degree 
distribution of both PlaNet and PhoNet roughly indicate a power-law behavior with exponential cut-offs towards the 
tail~\cite{acl:06,Mukherjee:06}. Furthermore, PhoNet is also characterized by a very high clustering coefficient. The topological 
properties of the two networks and the generative model explaining the emergence of these properties are summarized 
in~\cite{coling:08}. However, all the above properties are useful in characterizing the local patterns of the network and provide 
very little insight about its global structure.

\section{Spectral Analysis of PhoNet}\label{sec:ESAPh}

In this section we describe the procedure and results of the spectral analysis of PhoNet. We begin with computation of the spectrum of PhoNet. After the analysis of the spectrum, we systematically investigate the top few eigenvectors of PhoNet and attempt to characterize their linguistic significance. In the process, we also analyze the corresponding eigenvectors of LanGraph that helps us in characterizing the properties of languages.

\begin{figure*}
\begin{center}
\includegraphics[width=6in]{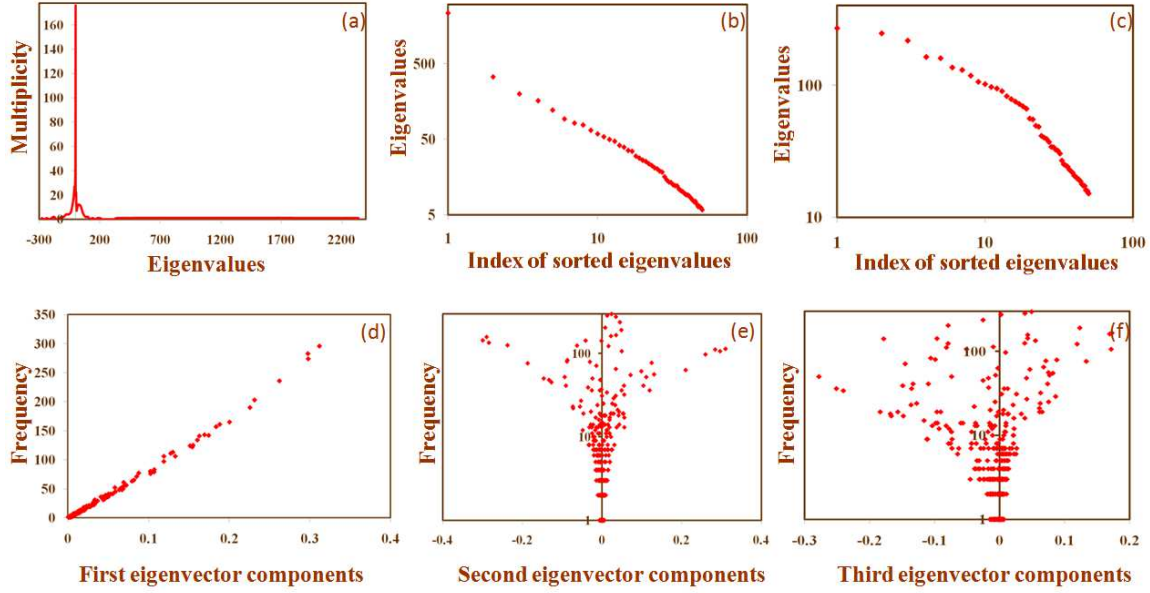}
\caption{Eigenvalues and eigenvectors of $\mathbf{B}$. (a) Binned distribution of the eigenvalues (bin size = 20) versus 
their multiplicities. (b) the top 50 (absolute) eigenvalues from 
the positive end of the spectrum and their ranks. (c) Same as (b) for the negative end of the spectrum. (d), (e) and (f) respectively represents the first, second and the third eigenvector components versus the occurrence frequency of the consonants.}\label{fig3}
\end{center}
\end{figure*}

\subsection{Spectrum of PhoNet}
 
Using a simple Matlab script we compute the spectrum (i.e., the list of eignevalues along with their multiplicities) of the matrix $\mathbf{B}$ corresponding to PhoNet.  Fig.~\ref{fig3}(a) shows the spectral plot, which has been obtained through {\em binning}\footnote{Binning is the process of dividing the entire range of a variable into smaller intervals and counting the number of observations within each bin or interval. In fixed binning, all the intervals are of the same size.} with a fixed bin size of 20. In order to have a better visualization of the spectrum, in Figs.~\ref{fig3}(b) and (c) we further plot the top 50 (absolute) eigenvalues from the two ends of the spectrum versus the index representing their sorted 
order in doubly-logarithmic scale. Some of the important observations that one can make from these results are as follows.

First, the major bulk of the eigenvalues are concentrated at around 0. This indicates that though the order of {\bf B} is 541 $\times$ 541, its numerical rank is quite low. Second, there are at least a few very large eigenvalues that dominate the entire spectrum. In fact, 89\% of the spectrum, 
or the square of the Frobenius norm, is occupied by the principal (i.e., the topmost) eigenvalue, 92\% is occupied by the first and the second eigenvalues taken together, while 93\% is occupied by the first three taken together. The individual contribution of the other eigenvalues to the spectrum is significantly lower than that of the top three. Third, the eigenvalues on either ends of the spectrum tend to decay gradually, mostly indicating a power-law behavior. The power-law exponents at the positive and the negative ends are -1.33 (the $R^2$ value of the fit is 0.98) and -0.88 ($R^2 \sim 0.92$) respectively.

The numerically low rank of PhoNet suggests that there are certain prototypical structures that frequently repeat themselves across the consonant inventories, thereby, increasing the number of 0 eigenvalues to a large extent. In other words, all the rows of the matrix $\mathbf{B}$ (i.e., the inventories) can be expressed as the linear combination of a few independent row vectors, also known as {\em factors}.

Furthermore, the fact that the principal eigenvalue constitutes 89\% of the Frobenius norm of the spectrum implies that there exist one very strong organizing principle which should be able to explain the basic structure of the inventories to a very good extent. Since the second and third eigenvalues are also significantly larger than the rest of the eigenvalues, one should expect two other organizing principles, which along with the basic principle, should be able to explain, (almost) completely, the structure of the inventories. In order to ``discover" these principles, we now focus our attention to the first three eigenvectors of PhoNet.

\subsection{The First Eigenvector of PhoNet}
Fig.~\ref{fig3}(d) shows the first eigenvector component for each consonant node versus its frequency of occurrence across the language inventories (i.e., its degree in PlaNet). The figure clearly indicates that the two are highly correlated ($r$ = 0.99), which in turn 
means that 89\% of the spectrum and hence, the organization of the consonant inventories, can be explained to a large 
extent by the occurrence frequency of the consonants. The question arises: Does this tell us something special about the structure of PhoNet or is it always the case for any symmetric matrix that the principal eigenvector will be highly correlated with the frequency? We assert that the former is true, and indeed, the high correlation between the principal eigenvector and the frequency indicates high ``proportionate co-occurrence" - a term which we will explain. 

To see this, consider the following $2n\times 2n$ matrix $\mathbf{X}$
\begin{displaymath}
\mathbf{X} = \left(\begin{array}{cccccc}
0 & M_1 & 0 & 0 & 0 & \dots \\
M_1 & 0 & 0 & 0 & 0 & \dots \\
0 & 0 & 0 & M_2 & 0 & \dots \\
0 & 0 & M_2 & 0 & 0 & \dots \\
\vdots & \vdots & \vdots & \vdots & \vdots & \ddots
\end{array}
\right)
\end{displaymath}
where $X_{i,i+1} = X_{i+1,i} = M_{(i+1)/2}$ for all odd $i$ and 0 elsewhere. Also, $M_1 > M_2 > \dots > M_n \ge 1$. Essentially, this matrix represents a graph which is a collection of $n$ disconnected edges, each having weights $M_1$, $M_2$, and so on. It is easy to see that the principal eigenvector of this matrix is $(1/\sqrt{2}, 1/\sqrt{2}, 0, 0, \dots, 0)^\top$, which of course is very different from the frequency vector: $(M_1, M_1, M_2, M_2, \dots, M_n, M_n)^\top$.  

At the other extreme, consider an $n \times n$ matrix $\mathbf{X}$ with $X_{i,j} = Cf_if_j$ for some vector $\mathbf{f} = (f_1, f_2, \dots f_n)^\top$ that represents the frequency of the nodes and a normalization constant $C$. This is what we refer to as "proportionate co-occurrence" because the extent of co-occurrence between the nodes $i$ and $j$ (which is $X_{i,j}$ or the weight of the edge between $i$ and $j$) is exactly proportionate to the frequencies of the two nodes. The principal eigenvector in this case is $\mathbf{f}$ itself, and thus, correlates perfectly with the frequencies. Unlike this hypothetical matrix $\mathbf{X}$, PhoNet has all 0 entries in the diagonal. Nevertheless, this perturbation, which is equivalent to subtracting $f_i^2$ from the $i^{th}$ diagonal, seems to be sufficiently small to preserve the ``proportionate co-occurrence" behavior of the adjacency matrix thereby resulting into a high correlation between the principal eigenvector component and the frequencies.  

On the other hand, to construct the Laplacian matrix, we would have subtracted $f_i\sum_{j=1}^n f_j$ from the $i^{th}$ diagonal entry, which is a much larger quantity than $f_i^2$. In fact, this operation would have completely destroyed the correlation between the frequency and the principal eigenvector component because the eigenvector corresponding to the smallest\footnote{The role played by the top eigenvalues and eigenvectors in the spectral analysis of the adjacency matrix is comparable to that of the smallest eigenvalues and the corresponding eigenvectors of the Laplacian matrix~\cite{chung:94}} eigenvalue of the Laplacian matrix is $[1, 1, \dots, 1]^\top$.

Since the first eigenvector of {\bf B} is perfectly correlated with the frequency of occurrence of the consonants across languages it is reasonable to argue that there is a universally observed innate preference towards certain consonants. This preference is often described through the linguistic concept of {\em markedness}, which in the context of phonology tells us that the substantive conditions that underlie 
the human capacity of speech production and perception renders certain consonants more favorable to be included in the inventory than some other consonants~\cite{Clements:04}. We observe that markedness plays a very important role in shaping the global structure of the consonant inventories. In fact, if we arrange the consonants in a non-increasing order of the first eigenvector components (which is equivalent to increasing order of statistical markedness), and compare the set of consonants present in an inventory of size $s$ with that of the first $s$ entries from this hierarchy, we find that the two are, on an average, more than 50\% similar. This figure is surprisingly high because, in spite of the fact that $\forall_s$ $s \ll \frac{541}{2}$, on an average $\frac{s}{2}$ consonants in an inventory are drawn from the first $s$ entries of the markedness hierarchy (a small set), whereas the rest $\frac{s}{2}$ are drawn from the remaining $(541-s)$ entries (a much larger set). 

The high degree of proportionate co-occurrence in PhoNet implied by this high correlation between the principal eigenvector and frequency further indicates that the innate preference towards certain phonemes is independent of the presence of other phonemes in the inventory of a language.

\subsection{The Second Eigenvector of PhoNet}
Fig.~\ref{fig3}(e) shows the second eigenvector component for each node versus their occurrence frequency. It is 
evident from the figure that the consonants have been clustered into three groups. Those that have a very low or a 
very high frequency club around 0 whereas, the medium frequency zone has clearly split into two parts. In order to 
investigate the basis for this split we carry out the following experiment.\\
{\em Experiment I}\\
(i) Remove all consonants whose frequency of occurrence across the inventories is very low ($<$ 5).\\
(ii) Denote the absolute maximum value of the positive component of the second eigenvector as $MAX_{+}$ and the 
absolute maximum value of the negative component as $MAX_{-}$. If the absolute value of a positive component is less 
than 15\% of $MAX_{+}$ then assign a {\em neutral} class to the corresponding consonant; else assign it a {\em 
positive} class. Denote the set of consonants in the {\em positive} class by $C_{+}$. Similarly, if the absolute value 
of a negative component is less than 15\% of $MAX_{-}$ then assign a {\em neutral} class to the corresponding 
consonant; else assign it a {\em negative} class. Denote the set of consonants in the {\em negative} class by 
$C_{-}$.\\
(iii) Using the above training set of the classified consonants (represented as boolean feature vectors) learn a 
decision tree (C4.5 algorithm~\cite{quinlan:93}) to determine the features that are responsible for the split of the 
medium frequency zone into the {\em negative} and the {\em positive} classes.

\begin{figure*}
\begin{center}
\includegraphics[width=5.5in]{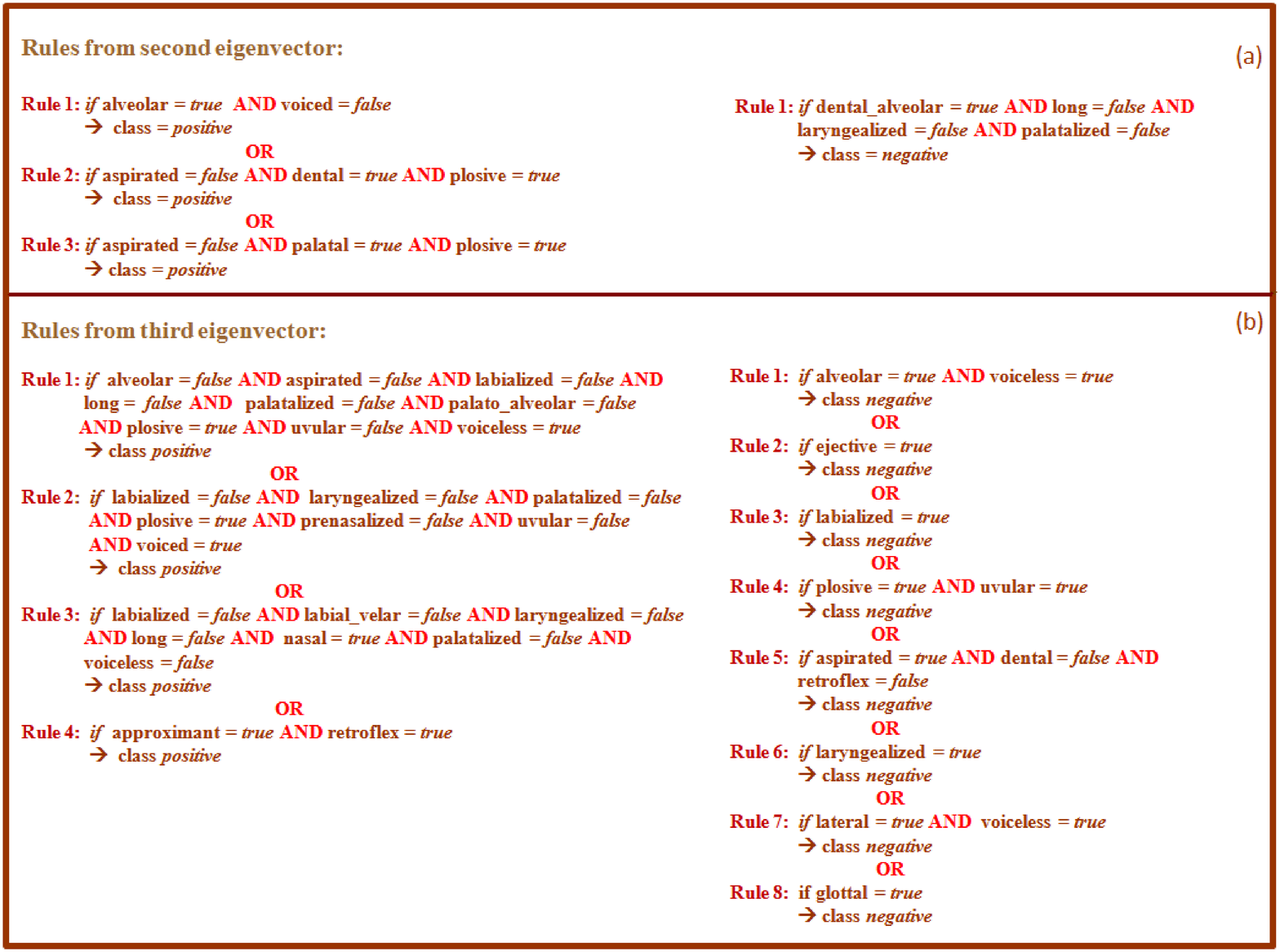}
\caption{Decision rules obtained from the study of (a) the second, and (b) the third eigenvectors. The classification 
errors for both (a) and (b) are less than 15\%.}\label{fig4}
\end{center}
\end{figure*}
   
Fig.~\ref{fig4}(a) shows the decision rules learnt from the above training set. It is clear from these rules that the 
split into $C_{-}$ and $C_{+}$ has taken place mainly based on whether the consonants have the combined 
``dental\_alveolar" feature ({\em negative} class) or the ``dental" and the ``alveolar" features separately ({\em 
positive} class). Such a combined feature is often termed {\em ambiguous} and its presence in a particular consonant 
$c$ of a language $l$ indicates that the speakers of $l$ are unable to make a distinction as to whether $c$ is 
articulated with the tongue against the upper teeth or the alveolar ridge. In contrast, if the features are present 
separately then the speakers are capable of making this distinction. In fact, through the following experiment, we 
find that the consonant inventories of almost all the languages in UPSID get classified based on whether they preserve 
this distinction or not.\\
{\em Experiment II}\\
(i) Construct {\bf B}$'$ = {\bf A}$^\textrm{T}${\bf A} -- {\bf D}$'$ (i.e., the adjacency matrix of LangGraph).\\ 
(ii) Compute the second eigenvector of {\bf B}$'$. Once again, the positive and the negative components split the 
languages into two distinct groups $L_{+}$ and $L_{-}$ respectively.\\
(iii) For each language $l \in L_{+}$ count the number of consonants in $C_{+}$ that occur in $l$. Sum up the counts 
for all the languages in $L_{+}$ and normalize this sum by $|L_{+}||C_{+}|$. Similarly, perform the same step for the 
pairs ($L_{+}$,$C_{-}$), ($L_{-}$,$C_{+}$) and ($L_{-}$,$C_{-}$).
   
From the above experiment, the values obtained for the pairs (i) ($L_{+}$,$C_{+}$), ($L_{+}$,$C_{-}$) are 0.35, 0.08 
respectively, and (ii) ($L_{-}$,$C_{+}$), ($L_{-}$,$C_{-}$) are 0.07, 0.32 respectively. This immediately implies that 
almost all the languages in $L_{+}$ preserve the dental/alveolar distinction while those in $L_{-}$ do not.

\subsection{The Third Eigenvector of PhoNet}
We next investigate the relationship between the third eigenvector components of {\bf B} and the occurrence frequency 
of the consonants (Fig.~\ref{fig3}(f)). The consonants are once again found to get clustered into three groups, though 
not as clearly as in the previous case. Therefore, in order to determine the basis of the split, we repeat experiments 
I and II. Fig.~\ref{fig4}(b) clearly indicates that in this case the consonants in $C_{+}$ lack the complex features 
that are considered difficult for articulation. On the other hand, the consonants in $C_{-}$ are mostly composed of 
such complex features. The values obtained for the pairs (i) ($L_{+}$,$C_{+}$), ($L_{+}$,$C_{-}$) are 0.34, 0.06 
respectively, and (ii) ($L_{-}$,$C_{+}$), ($L_{-}$,$C_{-}$) are 0.19, 0.18 respectively. This implies that while there 
is a prevalence of the consonants from $C_{+}$ in the languages of $L_{+}$, the consonants from $C_{-}$ are almost 
absent. However, there is an equal prevalence of the consonants from $C_{+}$ and $C_{-}$ in the languages of $L_{-}$. 
Therefore, it can be argued that the presence of the consonants from $C_{-}$ in a language can (phonologically) imply 
the presence of the consonants from $C_{+}$, but not vice versa. We do not find any such aforementioned pattern for 
the fourth and the higher eigenvector components.

\subsection{Control Experiment}
As a control experiment we generated a set of random inventories and carried out the experiments I and II on the 
adjacency matrix, {\bf B}$_\textrm{R}$, of the random version of PhoNet. We construct these inventories as follows. 
Let the frequency of occurrence for each consonant $c$ in UPSID be denoted by $f_c$. Let there be 317 bins each 
corresponding to a language in UPSID. $f_c$ bins are then chosen uniformly at random and the consonant $c$ is packed 
into these bins. Thus the consonant inventories of the 317 languages corresponding to the bins are generated. Note 
that this method of inventory construction leads to proportionate co-occurrence. 
Consequently, the first eigenvector components of {\bf B}$_\textrm{R}$ are highly correlated to the occurrence 
frequency of the consonants. However, the plots of the second and the third eigenvector components versus the 
occurrence frequency of the consonants indicate absolutely no pattern thereby, resulting in a large number of decision 
rules and very high classification errors (upto 50\%).
 
\section{Discussion and Conclusion}\label{sec:disc}

Are there any linguistic inferences that can be drawn from the results obtained through the study of the spectral plot 
and the eigenvectors of PhoNet? In fact, one can correlate several phonological theories to the aforementioned 
observations, which have been construed by the past researchers through very specific studies.

One of the most important problems in defining a feature-based classificatory system is to decide when a sound in one 
language is different from a similar sound in another language. According to Ladefoged~\shortcite{Ladefoged:05} 
``two sounds in different languages should be considered as distinct if we can point to a third language in which the 
same two sounds distinguish words". The dental versus alveolar distinction that we find to be highly instrumental in 
splitting the world's languages into two different groups (i.e., $L_{+}$ and $L_{-}$ obtained from the analysis of the 
second eigenvectors of {\bf B} and {\bf B}$'$) also has a strong classificatory basis. It may well be the case that 
certain categories of sounds like the dental and the alveolar sibilants are not sufficiently distinct to constitute a 
reliable linguistic contrast (see~\cite{Ladefoged:05} for reference). Nevertheless, by allowing the possibility for 
the dental versus alveolar distinction, one does not increase the complexity or introduce any redundancy in the 
classificatory system. This is because, such a distinction is prevalent in many other sounds, some of which are (a) 
nasals in Tamil~\cite{shan:72} and Malayalam~\cite{shan:72,Ladefoged:96}, (b) laterals in 
Albanian~\cite{Ladefoged:96}, and (c) stops in certain dialectal variations of Swahili~\cite{hayward:89}. Therefore, 
it is sensible to conclude that the two distinct groups $L_{+}$ and $L_{-}$ induced by our algorithm are true 
representatives of two important linguistic typologies.

The results obtained from the analysis of the third eigenvectors of {\bf B} and {\bf B}$'$ indicate that implicational 
universals also play a crucial role in determining linguistic typologies. The two typologies that are predominant in 
this case consist of (a) languages using only those sounds that have simple features (e.g., plosives), and (b) 
languages using sounds with complex features (e.g., lateral, ejectives, and fricatives) that automatically imply the 
presence of the sounds having simple features. The distinction between the simple and complex phonological features is 
a very common hypothesis underlying the implicational hierarchy and the corresponding typological 
classification~\cite{Clements:04}. In this context, Locke and Pearson~\shortcite{locke:92} remark that ``Infants 
heavily favor stop consonants over fricatives, and there are languages that have stops and no fricatives but no 
languages that exemplify the reverse pattern. [Such] `phonologically universal' patterns, which cut across languages 
and speakers are, in fact, the phonetic properties of Homo sapiens." (as quoted in~\cite{vallee:02}).   

Therefore, it turns out that the methodology presented here essentially facilitates the induction of linguistic 
typologies. Indeed, spectral analysis derives, in a unified way, the importance of these principles 
and at the same time quantifies their applicability in explaining the structural patterns observed across the 
inventories. In this context, there are at least two other novelties of this work. The first novelty is in the systematic study of the spectral plots (i.e., the distribution of the eigenvalues), which is in general rare for linguistic networks, although there have been quite a number 
of such studies in the domain of biological and social networks~\cite{farkas:01,mihail:03,Anirban:07}. The 
second novelty is in the fact that there is not much work in the complex network literature that investigates the nature of the eigenvectors and their interactions to infer the organizing principles of the system represented through the network. 

To summarize, spectral analysis of the complex network of speech sounds is able to provide a holistic as well as 
quantitative explanation of the organizing principles of the sound inventories. Although this natural mathematical 
technique has been heavily used in various other domains, we do not know of any work that uses spectral analysis for 
induction and understanding of linguistic typologies. This scheme for typology induction is not dependent on the 
specific data set used as long as it is representative of the real world. Thus, we believe that the scheme introduced 
here can be applied as a generic technique for typological classifications of phonological, syntactic and semantic 
networks; each of these are equally interesting from the perspective of understanding the structure and evolution of 
human language, and are topics of future research. 

\section*{Acknowledgement}
We would like to thank Kalika Bali for her valuable inputs towards the linguistic analysis.

\bibliographystyle{acl}
\bibliography{eacl09}

\end{document}